\documentclass{article}

\usepackage[preprint]{neurips_2026}           

\usepackage[utf8]{inputenc}
\usepackage[T1]{fontenc}
\usepackage{hyperref}
\usepackage{url}
\usepackage{microtype}
\usepackage{nicefrac}
\usepackage{enumitem}

\usepackage{amsmath, amssymb, amsthm}
\usepackage{mathtools}
\usepackage{bm}
\usepackage{algorithm}
\usepackage{algpseudocode}

\usepackage{booktabs}
\usepackage{multirow}
\usepackage{adjustbox}
\usepackage{array}

\usepackage{siunitx}
\usepackage[table]{xcolor}
\usepackage{colortbl}

\usepackage{graphicx}
\usepackage{subcaption}
\usepackage{caption}
\usepackage{float}
\usepackage{pifont}
\usepackage{hyphenat}

\usepackage{tikz}
\usepackage{pgfplots}
\pgfplotsset{compat=1.18}
\usetikzlibrary{
  arrows.meta, calc, decorations.pathreplacing, decorations.markings,
  fit, matrix, positioning, shadows, shapes.geometric, shapes.misc,
  backgrounds, patterns, fadings
}

\definecolor{lightgray}{gray}{0.92}
\definecolor{headerblue}{RGB}{240,245,255}
\definecolor{deltagreen}{RGB}{0,128,0}
\definecolor{deltared}{RGB}{180,0,0}
\definecolor{deltarow}{gray}{0.93}
\definecolor{improvement}{rgb}{0.0,0.42,0.10}
\definecolor{ourblue}{RGB}{220,234,250}
\definecolor{bettergreen}{RGB}{0,130,60}
\definecolor{worsered}{RGB}{180,0,0}
\definecolor{PrismBlue}{RGB}{30,100,200}
\definecolor{PrismRed}{RGB}{210,50,50}
\definecolor{StageGray}{RGB}{240,242,248}
\definecolor{StageGrayDark}{RGB}{200,210,230}
\definecolor{NodeFill}{RGB}{245,248,255}
\definecolor{NodeBorder}{RGB}{70,120,200}
\definecolor{HeaderBlue}{RGB}{20,60,140}

\newcommand{\dmodel}{d_{\text{model}}}
\newcommand{\dhead}{d_{\mathrm{head}}}

\title{Prism Transformer: Progressive Head Schedules \\for Hierarchical Attention Processing}

\author{%
  Shubham Aggarwal \\
  \texttt{Shubham\_agg@alumni.iitm.ac.in}
}

\begin{document}
\maketitle

\begin{abstract}
Multi-head attention conventionally partitions the hidden dimension equally across all heads at every layer, enforcing an identical representational subspace dimension ($d_h = d_{\text{model}}/h$) throughout the model's depth. In this work, we identify this uniform allocation as a fundamental structural bottleneck: due to their restricted dimensional space, early-layer heads are unable to faithfully capture complex, high-dimensional contextual patterns. To resolve this, we introduce the Prism Transformer, a novel architectural paradigm that replaces the static, uniform head configuration with a progressive head schedule. By monotonically increasing the head count across layers, the Prism Transformer naturally establishes a local-to-global representational hierarchy: early layers leverage fewer, exceptionally wide heads to capture complex, local compositional patterns, while deep layers deploy many, narrow heads to decompose these patterns into specialized linguistic features. Crucially, this structural shift is parameter-neutral, compute-neutral, and introduces zero training or inference overhead, preserving identical weight matrices and FLOP budgets as the standard Transformer. Across three model scales (124M, 354M, and 757M), the Prism Transformer consistently outperforms uniform baselines, achieving consistent reductions in validation loss alongside consistent gains on downstream zero-shot benchmarks (including PIQA, HellaSwag, ARC-Easy, and WinoGrande). Our findings demonstrate that non-uniform subspace allocation unlocks latent capacity within the standard Transformer budget, enabling more effective use of model capacity.
\end{abstract}

\section{Introduction}
\label{sec:intro}

The multi-head attention (MHA) mechanism~\cite{vaswani2023attentionneed} is the
defining component of the Transformer architecture.  By projecting queries,
keys, and values into $h$ independent subspaces of dimension
$d_h = \dmodel / h$, MHA enables the model to simultaneously attend to multiple distinct
representational patterns. This division of the hidden dimension has been held constant across all layers since the original Transformer: every layer in a standard decoder uses the identical number of heads, and therefore operates within the exact same per-head dimension.

This uniform allocation embeds a strong, implicit assumption about representation processing: that all layers across a network's depth benefit equally from the same granularity of attention subspaces. We argue that this assumption introduces a fundamental architectural mismatch between what early layers require and what a uniform schedule forces them to execute.

\paragraph{The problem with uniform allocation}
Early transformer layers are responsible for integrating raw token embeddings into high-level representations that capture complex, local compositional semantic structures. Encoding these intricate local patterns faithfully requires substantial representational capacity per subspace. Under a standard uniform configuration (e.g., $h=12$ heads over $d_{model}=768$), each early-layer head is restricted to a narrow $d_{h}=64$ dimensions. Due to this severely restricted dimensional space, individual heads are forced to thinly distribute a small subspace across positions, rendering them physically ill-equipped to resolve complex, high-dimensional local contextual features.

As representations progress through the network, their structural needs evolve. Mid-network layers serve as the primary engine for global contextual integration, aggregating these dense local features across long-range sequence dependencies. Finally, late transformer layers shift away from broad integration toward specialized feature decomposition, extracting and isolating specific fine-grained syntactic, semantic, or task-oriented signals. This downstream refinement objective is naturally well-suited to a high density of narrow, focused subspaces operating in parallel. By forcing every stage, including local composition, global integration, and fine-grained refinement, into identical structural configurations, the conventional uniform schedule ignores the natural local-to-global trajectory of the network.

\paragraph{The Prism Transformer.}
We propose to resolve this structural bottleneck by replacing the rigid, uniform configuration with a progressive head schedule: a non-decreasing sequence $(h_{1}, h_{2}, \dots, h_{L})$ where $h_{l}$ denotes the head count of layer $l$. Early layers utilize a small head count, granting each individual head an exceptionally wide subspace ($d_{h}^{(l)} = d_{\text{model}}/h_{l}$ is large when $h_{l}$ is small). The head count then increases monotonically across the network's depth, converging to the standard baseline configuration in later layers.

The shape of this schedule traces a decreasing trajectory in $d_{h}$ across depth; forming a narrowing prism through which representations flow from local-to-global.
This design yields two critical structural properties. First, it is parameter-neutral: because the MHA projection matrices ($W_Q, W_K, W_V, W_O$) maintain their standard $\mathbb{R}^{d_{\text{model}} \times d_{\text{model}}}$ shapes, altering the number of slices does not alter the total parameter count. Second, it is compute-neutral: the dominant attention FLOPs remain mathematically invariant to head count. The Prism Transformer thus unlocks latent representational capacity completely for free.

\begin{figure}[t]
    \centering
    \includegraphics[width=\linewidth]{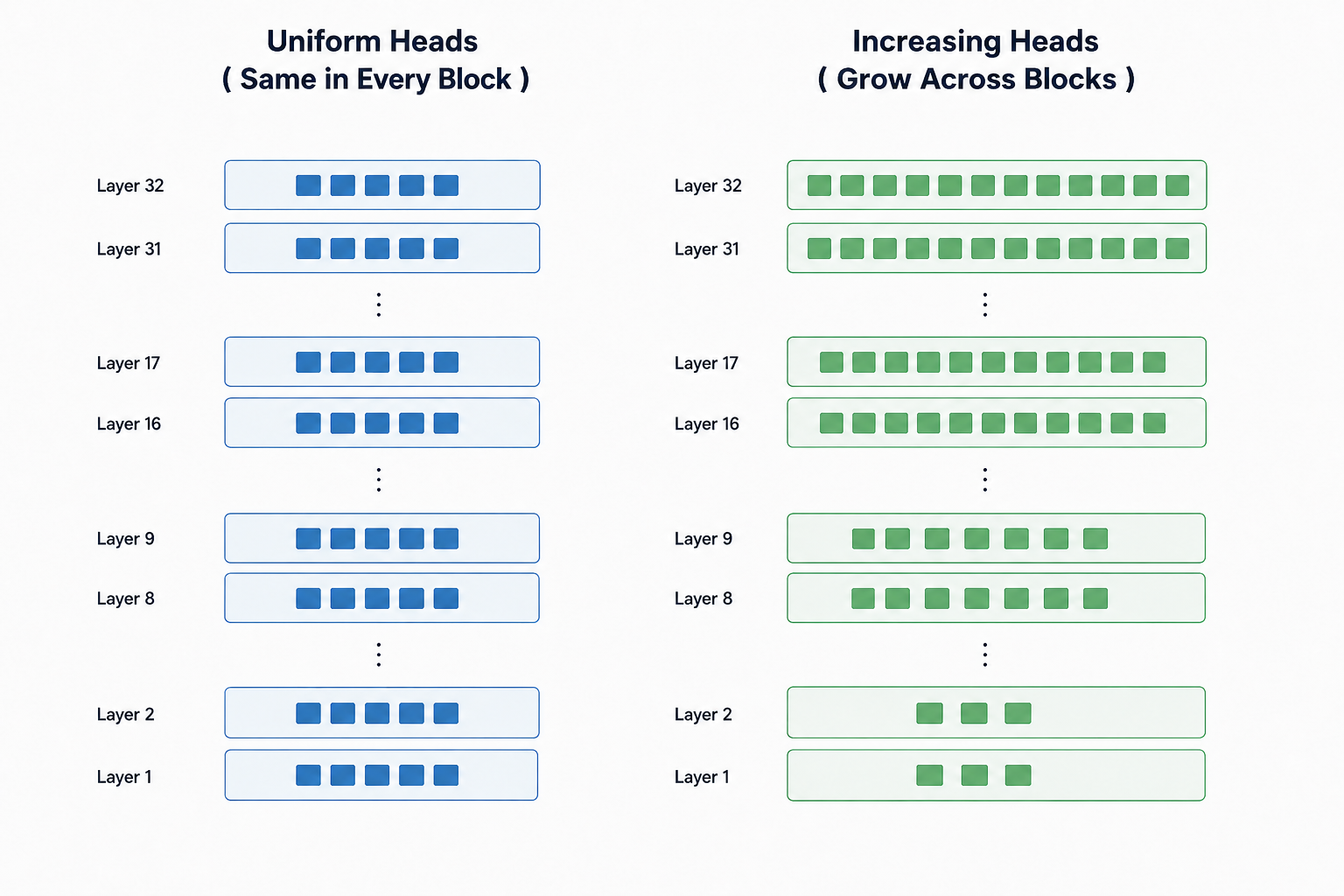}
    \caption{%
        \textbf{Head schedule visualization}
         Comparison of attention head allocations across layers. (Left) Standard baseline featuring uniform heads (a flat schedule across all blocks). (Right) The Prism Transformer featuring an increasing head schedule (a progressive staircase growth across blocks). By expanding the head count in deeper layers, the Prism Transformer implicitly creates a monotonically decreasing per-head dimension ($d_h$) trajectory.
    }
    \label{fig:head_schedule_vis}
\end{figure}

Figure~\ref{fig:head_schedule_vis} illustrates the contrast between a uniform and
a Prism schedule.

\paragraph{Contributions} Our primary contributions are as follows:
\begin{itemize}[leftmargin=*]
    \item \textbf{Prism Transformer:} We propose a progressive head schedule for decoder-only Transformers that establishes a robust local-to-global representational hierarchy at zero parameter or compute overhead.

    \item \textbf{Consistent Scale-Invariant Gains:} Across three model scales
    (124M, 354M, 757M), the Prism Transformer achieves lower validation
    loss than the uniform baselines at identical training compute.
    (Table~\ref{tab:combined_results}).

    \item \textbf{Mechanistic Attention Analysis:} We conduct a detailed per-layer attention distance analysis, empirically demonstrating that the progressive schedule restructures the network's attention profile. It encourages tight, highly local semantic aggregation in early wide-head layers and shifts broad, global integration to the mid-network layers where the schedule completes (Section ~\ref{sec:analysis}).

    \item \textbf{Downstream Benchmark Parity and Improvements:}
    Evaluation across zero-shot language benchmarks (PIQA, HellaSwag, ARC-Easy, and WinoGrande) confirms that our progressive schedule preserves or explicitly improves downstream task-level accuracy. (Table~\ref{tab:benchmarks}, Figure~\ref{fig:benchmark_results}).

    \item \textbf{Head schedule design principles:} Through systematic structural ablations, we isolate the specific geometric criteria, such as dimension transition smoothness and block consolidation, that govern schedule effectiveness. We formalize these findings into a compact set of transferable design rules that generalize consistently across all evaluated parameter scales (Section~\ref{sec:method}).
\end{itemize}

\section{The Prism Transformer}
\label{sec:method}

\subsection{Mathematical Formulation of Progressive Head Schedules}
\label{subsec:phs}

Let an $L$-layer Transformer possess a constant model dimension $d_{\text{model}}$. A progressive head schedule is defined as a non-decreasing integer sequence $\mathcal{S} = (h_1, h_2, \dots, h_L)$ where $h_l$ denotes the allocation of attention heads at layer $l$, subject to the following structural constraints:  

\begin{enumerate}[label=(\roman*), leftmargin=*]
    \item Divisibility: $h_l \mid d_{\text{model}}$ for all $l \in \{1, \dots, L\}$ \quad (head count divides model dimension),
    \item Monotonicity: $h_1 \le h_2 \le \cdots \le h_L$ \quad (monotonically non-decreasing),
    \item Boundary Convergence: $h_L = h_{\text{base}}$ \quad (final layers match the baseline head count),
\end{enumerate}

\subsection{Empirical Design Guidelines}
While the constraints in Section 2.1 define the space of valid schedules, optimization within this space requires tuning. Through systematic structural ablations on candidate shapes (detailed in Appendix ~\ref{sec:appendix_ablations}), we formalize two vital design principles that dictate schedule effectiveness:

\begin{enumerate}[label=(\roman*), leftmargin=*]
    \item Staircase Smoothness: Abrupt, single-step transitions degrade performance. Optimal schedules utilize localized, multi-layer consolidation phases (e.g., maintaining a specific head dimension for at least 2 to 4 consecutive layers) to stabilize representations before altering granularity.
    \item Baseline Preservation Phase: To ensure robust semantic convergence, at least half of the network's total layers ($L/2$) must be dedicated to the baseline head count $h_{\text{base}}$.

\end{enumerate}

\subsection{Hardware Alignment \& Complexity}

A primary advantage of the Prism Transformer is its zero-overhead integration into modern compute clusters. Altering head counts does not touch the parameter space: because the projection matrices ($W_Q, W_K, W_V, W_O \in \mathbb{R}^{d_{\text{model}} \times d_{\text{model}}}$) maintain their standard uniform shapes, total parameters remain strictly invariant.

Furthermore, we impose a hardware layout preference when selecting among valid schedules. By choosing configurations where the early-stage head counts yield dimensions that are powers of two ($d_h^{(l)} \in \{256, 128\}$) , the resulting attention slices align natively with the tile boundaries of GPU Tensor Cores. This avoids execution memory strides and unaligned tensor splits, guaranteeing that our representational gains introduce zero practical latency penalties during standard training or inference passes.

\subsection{Parameter and Compute Neutrality}
\label{subsec:neutrality}

\paragraph{Parameter Invariance}
The parameter budget of a standard multi-head attention block at layer $l$ is determined entirely by the projection operators for queries, keys, values, and output mapping: $W_Q, W_K, W_V, W_O \in \mathbb{R}^{d_{\text{model}} \times d_{\text{model}}}$. The total parameter count for a single layer's MHA module is formalized as:

$$N_{\text{params}}^{(l)} = 4d_{\text{model}}^2 + 4d_{\text{model}} \cdot \mathbb{I}_{\text{bias}}$$

where $\mathbb{I}_{\text{bias}} \in \{0, 1\}$ represents an indicator variable for the presence of a bias vector. Because this allocation depends strictly on the global model dimension $d_{\text{model}}$, it remains fundamentally invariant to the layer-specific head count $h_l$. Thus, the Prism Transformer achieves its architectural re-balancing with exactly zero parameter inflation or structural footprint modifications.

\paragraph{Computational Complexity}
The primary floating-point operations (FLOPs) per token step in a multi-head attention layer scale according to three distinct operations. Let $T$ denote the sequence context length. The computational breakdown per layer $l$ proceeds as follows:

\begin{enumerate}[label=(\roman*), leftmargin=*]
    \item Dense Projections: The linear mappings to obtain $Q, K,$ and $V$ representations require $\mathcal{O}(T \cdot d_{\text{model}}^2)$ FLOPs.
    \item Attention Matrix Computation: Computing the inner-product attention matrix and applying softmax requires scaling by the number of heads multiplied by their respective subspace dimensions:

$$\mathcal{O}\left(T^2 \cdot h_l \cdot d_h^{(l)}\right) = \mathcal{O}\left(T^2 \cdot d_{\text{model}}\right)$$

    \item Output Linear Alignment: The final projection matrix $W_O$ requires $\mathcal{O}(T \cdot d_{\text{model}}^2)$ FLOPs.

\end{enumerate}

Because none of these algorithmic terms depend on the choice of $h_l$, the Prism Transformer is strictly compute-neutral in theoretical FLOP allocation.

\paragraph{Hardware alignment}
In the Prism Transformer, we formulate the progressive head schedule in such a way that the early-stage head configurations yield dimensions that are powers of two ($d_h^{(l)} \in \{256, 128\}$). This structural choice natively preserves hardware tile and memory alignment across GPU Tensor Cores, completely avoiding unaligned tensor splits or memory strides. Consequently, the Prism Transformer naturally matches the raw throughput and wall-clock training speeds of standard uniform models, delivering its architectural representational advantage at zero computational cost. (Table ~\ref{tab:combined_results}).

\section{Experiments}
\label{sec:experiments}

\subsection{Experimental Setup}
\label{subsec:setup}

\paragraph{Hardware and Core Implementation}
All training runs are executed on a distributed cluster consisting of $8\times$ NVIDIA H100 (80GB SXM5) GPUs interconnected via NVLink. Models are built in PyTorch by extending the NanoGPT framework~\cite{karpathy_nanogpt_2022} with Rotary Positional Embeddings
(RoPE)~\cite{su2023roformerenhancedtransformerrotary} and SwiGLU
activations~\cite{shazeer2020gluvariantsimprovetransformer}.  Architectural adjustments are completely isolated to the attention layer split dimensions; all other macro-parameters (e.g., hidden states, layer depths, optimizer choices) remain identical across baseline and experimental setup

\paragraph{Dataset and Tokenization}
Models are pre-trained on the FineWeb dataset~\cite{penedo2024the}, tokenized with the GPT-2
BPE tokenizer.  Following the Chinchilla scaling
law~\cite{hoffmann2022trainingcomputeoptimallargelanguage}, we train each model
for 25 to 30 tokens per parameter, slightly exceeding the compute-optimal ratio to
ensure all variants reach sufficient convergence for a robust comparison.

\paragraph{Training Configuration}
Training uses mixed-precision \texttt{bfloat16} under Distributed Data Parallel
(DDP) with a fixed context length of 1024 tokens.  The AdamW optimizer is
configured with $\beta_1 = 0.9$, $\beta_2 = 0.95$, $\epsilon = 10^{-8}$, weight
decay $0.1$, and gradient clipping at $1.0$.  A cosine decay learning rate schedule
with linear warmup over the first 2.5\% of training tokens is used throughout.
Architectural specifications and hyperparameters are summarized in
Table~\ref{tab:model_configs}.

\begin{table}[t]
\centering
\caption{Model configurations and training hyperparameters. $h_{\text{base}}$
denotes the baseline (uniform) head count; the Prism schedule uses
$h_{\text{base}}$ only in later layers (see Appendix~\ref{app:schedules}).
All other architectural hyperparameters are identical across the baseline and
Prism variants.}
\label{tab:model_configs}
\setlength{\tabcolsep}{6pt}
\renewcommand{\arraystretch}{1.15}
\begin{adjustbox}{max width=\linewidth}
\begin{tabular}{lccccccc}
\toprule
\textbf{Model} & $n_{\text{params}}$ & $n_{\text{layers}}$ & $\dmodel$
  & $h_{\text{base}}$ & Batch Size & LR & Training Tokens \\
\midrule
Small  & 124M & 12 & 768  & 12 & 0.25M & $6.0\times10^{-4}$ & 2.88B \\
Medium & 354M & 24 & 1024 & 16 & 0.25M & $3.0\times10^{-4}$ & 10.49B \\
Large  & 757M & 24 & 1536 & 16 & 0.50M & $2.5\times10^{-4}$ & 20.97B \\
\bottomrule
\end{tabular}
\end{adjustbox}
\end{table}

\subsection{Training Results}
\label{subsec:main_results}

To evaluate the fundamental architectural impact of progressive head schedules, we compare the pre-training performance of the Prism Transformer against uniform baseline configurations across three distinct parameter scales. To ensure our empirical conclusions are structurally sound and independent of optimization initialization noise, all configurations are trained over three independent random seeds. We report the empirical mean and sample standard deviation ($\mu \pm \sigma$) calculated at the terminal training checkpoints.

\begin{table*}[t]
\centering
\caption{Validation performance on FineWeb and hardware execution benchmarks on an $8\times$ NVIDIA H100 GPU cluster. Validation metrics are reported as mean $\pm$ standard deviation across three independent runs. Throughput and wall-clock measurements are empirical means across three training seeds. Bold indicates best-performing configuration.}
\label{tab:combined_results}
\resizebox{\textwidth}{!}{
\begin{tabular}{llccc}
\toprule
\textbf{Params} & \textbf{Model} & \textbf{Val Loss} & \textbf{Tokens/sec} & \textbf{Wall-clock (hrs)} \\
\midrule

\multirow{2}{*}{124M}
& Baseline Uniform      & $3.3748 \pm 0.0027$ & 2875K & 0.31 \\
& \textbf{Prism Transformer} & $\mathbf{3.3604 \pm 0.0010}$ & 2875K & 0.31 \\
\midrule

\multirow{2}{*}{354M}
& Baseline Uniform      & $2.9944 \pm 0.0015$ & 1025K & 3.03 \\
& \textbf{Prism Transformer} & $\mathbf{2.9873 \pm 0.0020}$ & 1025K & 3.03 \\
\midrule

\multirow{2}{*}{757M}
& Baseline Uniform      & $2.8106 \pm 0.0002$ & 656K & 9.09 \\
& \textbf{Prism Transformer} & $\mathbf{2.8022 \pm 0.0013}$ & 656K & 9.09 \\
\bottomrule

\end{tabular}
}
\end{table*}



As shown in Table~\ref{tab:combined_results},
the Prism Transformer consistently achieves lower validation loss across all
scales. Importantly, this
improvement is achieved purely through schedule redistribution; the model has exactly the same training token budgets, parameter counts, and total FLOP allocations as the standard architecture, but uses it more efficiently by aligning the representational granularity with layer depth.

\subsection{Downstream Benchmark Evaluation}
\label{subsec:benchmarks}

We evaluated our final model checkpoints on a broad suite of zero-shot downstream tasks using the LM Evaluation Harness~\cite{eval-harness}. To ensure statistical validity, we evaluated all three independent training seeds for both the uniform baselines and the Prism configurations across multiple foundational benchmarks: PIQA~\cite{piqa}, HellaSwag~\cite{hellaswag}, ARC-Easy~\cite{arc}, WinoGrande~\cite{sakaguchi2019winograndeadversarialwinogradschema}, BLiMP~\cite{blimp}, and WikiText~\cite{merity2016pointer}. We report the empirical mean accuracy ($\mu$) across these runs, with full tabulated results provided in Table~\ref{tab:benchmarks}.

\begin{figure}[t]
    \centering
    \includegraphics[width=\linewidth]{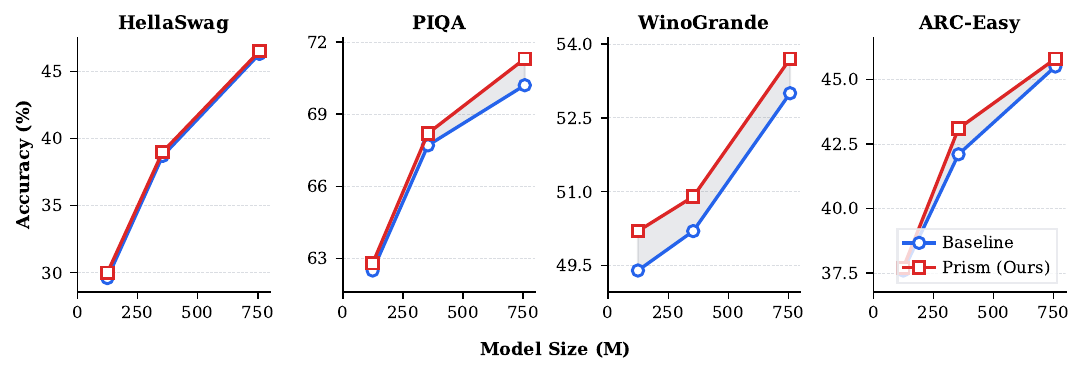}
    \caption{\textbf{Zero-shot benchmark performance across model scales.}
    All comparisons are evaluated at identical token milestones. The Prism Transformer
    achieves comparable or superior accuracy to the baseline across all tasks.}
    \label{fig:benchmark_results}
\end{figure}

As shown in Figure~\ref{fig:benchmark_results}, the Prism Transformer consistently preserves or explicitly improves downstream task-level accuracy across all evaluated model sizes. This demonstrates that expanding the early-layer subspace width provides the network with the necessary representational bandwidth to encode more robust, generalizable features that directly benefit contextual knowledge retrieval and structural common-sense reasoning.

\subsection{Hardware and System Training Throughput}
\label{subsec:efficiency}

To verify the system efficiency of the Prism Transformer on modern hardware clusters, we record wall-clock training durations and raw system execution throughput metrics across our distributed environments. We evaluate device metrics across all three independent training seeds to isolate hardware variations, reporting the empirical mean execution speed (tokens per second) and total accumulated wall-clock runtime.

The hardware benchmarks presented in Table~\ref{tab:combined_results} demonstrate that our progressive subspace allocation achieves absolute operational parity with standard configurations. By enforcing a power-of-two constraint on early-layer head configurations ($d_h^{(l)} \in \{256, 128\}$), the Prism Transformer maintains native tensor tile alignment with GPU Tensor Cores. This intentional engineering alignment completely prevents unaligned tensor splits or memory layout strides, allowing the progressive architecture to identically match the raw execution throughput and total wall-clock runtimes of uniform models across all evaluated parameter scales.

Furthermore, our runs reveal an additional optimization advantage: Prism Transformer models reach their minimal validation loss milestones consistently earlier in training than uniform baselines, resulting in a practical reduction in the total computational footprint. This behavior suggests that matching attention subspace widths to the network's natural processing hierarchy yields faster convergence dynamics alongside improved representation accuracy, establishing this method as a highly practical alternative for compute-bounded LLM development.

\subsection{Scaling Experiments}
To evaluate the robustness and predictability of progressive head schedules under compute and parameter scaling, we evaluate the Prism Transformer against uniform baselines across three distinct model capacities: \textbf{Small (124M)}, \textbf{Medium (354M)}, and \textbf{Large (757M)}. All model pairs are trained from scratch under identical hyperparameter configurations, dataset mixtures, and token budgets to isolate the structural impact of the head schedule (see Appendix~\ref{app:schedules} for exact schedule details).

\begin{figure}[t]
    \centering
    \includegraphics[width=\linewidth]{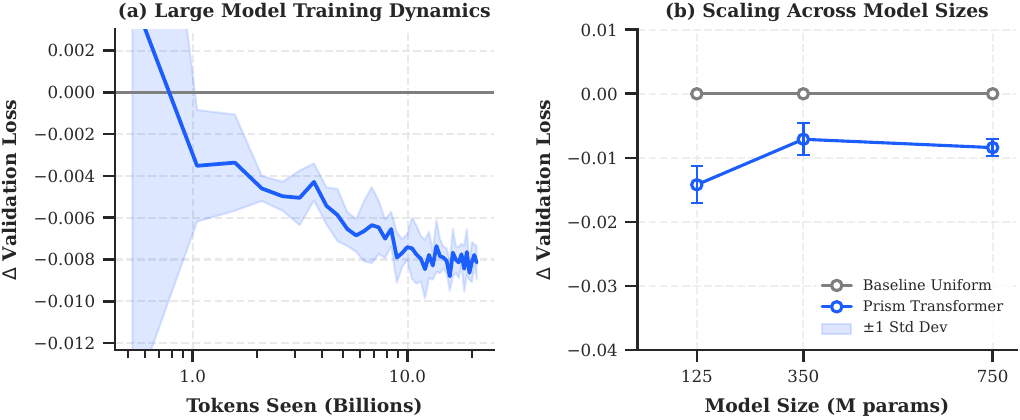}
    \caption{\textbf{Scaling properties of Prism Transformer compared to Baseline Uniform.}
    (a) Training Dynamics. Solid line depicts the mean validation loss gap across three random seeds for the large model, with the shaded region denoting ±1 standard deviation. A negative gap indicates Prism outperforms the baseline. (b) \textbf{Model Scale Generalization.} Mean validation loss gap at convergence across model sizes (124M, 354M, 757M parameters), with error bars denoting ±1 standard deviation propagated across seeds}
    \label{fig:scaling_curves}
\end{figure}

The pretraining scaling trajectories illustrated in Figure~\ref{fig:scaling_curves} yield several key insights:

\begin{itemize}
    \item \textbf{Consistent Scaling Advantage:} Across all three parameter scales, the Prism Transformer consistently achieves a lower validation cross-entropy loss compared to its uniform baseline counter-part. Crucially, as shown in Figure~\ref{fig:scaling_curves} (right), this performance gap does not narrow or diminish as model capacity increases from 124M to 757M.
    
    \item \textbf{Widening Training Gap:} As tracked in the large model training dynamics in Figure~\ref{fig:scaling_curves} (left), the validation loss delta exhibits distinct phase behaviors over the course of training. Following an initial high-variance regime below 1.0B tokens, the performance advantage of the Prism Transformer rapidly expands between 1.0B and 10.0B tokens, where the negative gap widens significantly. In the final phase of training (beyond 10.0B tokens), this gap plateaus and stabilizes near $-0.008$, while the corresponding variance band ($\pm1$ Std Dev) contracts substantially.
    
\end{itemize}


\section{Analysis}
\label{sec:analysis}

\subsection{Attention Distance Reveals a Sharpened Local-to-Global Gradient}
\label{subsec:attention_distance}

To probe how the Prism schedule systematically reorganizes sequence context
processing, we track the mean attention distance at every layer for both
architectures.
For layer $l$ and head $i$, the attention-weighted mean token distance of a
query at position $t$ is $\sum_{s=1}^{t} A^{(l,i)}_{t,s}\,|t-s|$, where
$A^{(l,i)}_{t,s} = \operatorname{Softmax}\!\left(q^{(l,i)}_t
{K^{(l,i)}}^{\!\top}/\sqrt{\dhead}\right)_{\!s}$
are the causal attention weights ($A^{(l,i)}_{t,s}=0$ for $s>t$ by
construction).
We report the per-layer distance metric $\mathcal{D}_l$ by averaging across
heads, query positions in the last 50\% of the context window, and evaluation
sequences:
\begin{equation}
\mathcal{D}_l =
    \frac{1}{h_l} \sum_{i=1}^{h_l}
    \frac{1}{|\mathcal{Q}|} \sum_{t \in \mathcal{Q}}
    \mathbb{E}_{x}\!\left[
        \sum_{s=1}^{t} A^{(l,i)}_{t,s}\,|t - s|
    \right],
\quad \mathcal{Q} = \{\lfloor T/2  \rfloor +1, \ldots, T\},
\label{eq:attn_dist}
\end{equation}
where $\mathbb{E}_{x}$ denotes the empirical expectation over validation
sequences. Restricting $\mathcal{Q}$ to positions $t \geq \lfloor T/2
\rfloor +1 = 513$ (for $T=1024$) ensures every query has at least 512 tokens
of available context, eliminating the artefact whereby early tokens are
forced to attend locally due to a short causal history.
Because the distance functional is linear in the attention weights, the head
average in Eq.~\eqref{eq:attn_dist} is mathematically identical to computing
the distance on a single head-averaged attention
map~\cite{vig2019analyzingstructureattentiontransformer,
raghu2022visiontransformerslikeconvolutional}, so $\mathcal{D}_l$ is
independent of both $h_l$ and $\dhead$, enabling a direct comparison across
schedules with different head counts. All values are in absolute token units,
averaged over three independent seeds ($n=3$).

Figure~\ref{fig:attn_distance} shows the resulting profiles across all three
model scales. In the early layers, where the Prism schedule allocates wider
subspaces, $\mathcal{D}_l$ is substantially lower than the baseline
($\Delta\mathcal{D}_l < 0$), indicating sharper local aggregation. This
reverses in the mid-network layers, where $\Delta\mathcal{D}_l$ peaks
positively, precisely at the point where the progressive ramp completes.
The fact that the sign change aligns with the schedule boundary and not
with any other structural transition confirms that this reorganization is
driven by the head allocation rather than being an optimisation artefact.
Both models converge to identical attention patterns in the final layers.
We emphasize that this analysis provides a mechanistic signature of the
Prism schedule's effect; the primary performance justification (including matched parameters, identical wall-clock throughput, and superior downstream task accuracy) remains anchored in our core empirical evaluations established in Section ~\ref{sec:experiments}.

\subsection{Why Does the Progressive Schedule Win?}

Combining the attention-distance analysis (Section~\ref{subsec:attention_distance}) with the progressive head trajectory (Section~\ref{sec:method}), we
offer the following interpretive account of the Prism Transformer's advantage.
We stress that our core empirical claim rests on the measured reorganization of attention distance; the causal narrative below provides a structural interpretation of these findings.

\begin{enumerate}[leftmargin=*, label=\textbf{(\arabic*)}]
    \item \textbf{Wide early heads build rich \emph{local} features}
    By allocating a small number of heads in the first layers, each individual head operates within an exceptionally wide 256-dimensional subspace ($d_h = 256$) while focusing tightly on immediate token neighborhoods (reflected by the negative early $\Delta\mathcal{D}_l$ in Figure ~\ref{fig:attn_distance}). This lets early layers construct highly expressive representations of local context, rather than thinly distributing a small
    subspace across many positions.

    \item \textbf{The mid-network head influx is dedicated to global integration}
   As the schedule scales up the head count across the middle layers, consequently narrowing individual subspace dimensions, these newly introduced heads are deployed for longer-range mixing. This is confirmed by the positive peak in $\Delta\mathcal{D}_l$, which directly aligns with the completion of the progressive phase. The steady progression of the staircase allows representations to stabilize at each head-count tier before transitioning.

    \item \textbf{Late layers refine well-integrated representations}
    By the time the head allocation reaches the baseline configuration, where heads operate at the standard uniform head dimension, the attention heads are processing representations whose local structures have already been established and globally integrated. The convergence of the attention distance curves in the final layers suggests that both models execute a comparable style of fine-grained refinement at this stage.
\end{enumerate}

This account predicts that the advantage is structural rather than an optimization artifact, meaning it should persist throughout the entire training trajectory. The token scaling curves in Section 3.2 (Figure ~\ref{fig:scaling_curves}) strongly validate this hypothesis: following the initial warmup period, the Prism Transformer maintains a lower validation loss at every token milestone.

\section{Comparison with Related Work}

\paragraph{Head Redundancy and Capacity} 
Standard multi-head attention layouts suffer from severe functional redundancy, where a large majority of attention heads can be pruned post-training with minimal impact on accuracy~\cite{michel2019sixteenheadsreallybetter, voita2019analyzingmultiheadselfattentionspecialized}. Furthermore, scaling head size inversely with head count introduces a low-rank bottleneck that restricts individual head expressivity~\cite{bhojanapalli2020lowrankbottleneckmultiheadattention}. While prior works address these inefficiencies via post-hoc compression or static adjustments, the Prism Transformer redesigns the head allocation schedule from the outset. By preventing over-partitioning in early layers, Prism natively mitigates redundancy and rank deficiencies before they form.

\paragraph{Depth and Layer Specialization} 
Probing studies show that Transformer layers naturally form a hierarchical processing pipeline, moving from early surface-level syntax to deep semantic abstractions~\cite{tenney2019bertrediscoversclassicalnlp}. Prior works exploit this specialization using structured dropout (LayerDrop)~\cite{fan2019reducingtransformerdepthdemand}, early-exit routing, or variable-width parameter allocations across depth~\cite{wu2026variablewidthtransformers}. Unlike these methods, which alter either active layer counts or total parameters per layer, Prism remains strictly compute-neutral. All layers remain active, but we systematically change how each layer geometrically segments its hidden dimensions across the network depth.

\paragraph{Inference-Optimized Head Layouts}
Multi-Query Attention (MQA)~\cite{shazeer2019fast} and Grouped-Query Attention (GQA)~\cite{ainslie2023gqa} modify head layouts to alleviate the Key-Value (KV) cache memory bottleneck during inference. They alter the \textit{ratio} of Query to KV projections within a static layer footprint, keeping head dimensions uniform across depth. Conversely, Prism targets representational expressivity across depth. It maintains symmetric Q, K, and V counts but dynamically scales head dimensions across layers to align with the processing hierarchy. Consequently, Prism is fully orthogonal to and composable with GQA.

\section{Conclusion}

We introduced the Prism Transformer, a simple modification to the standard multi-head attention mechanism that replaces the uniform head schedule with a progressive one. By starting with fewer, wider heads in early layers and gradually increasing the head count across depth, the Prism Transformer creates a local-to-global representational hierarchy that naturally aligns with the functional demands at each layer index. 

The modification is entirely parameter-neutral and compute-neutral by construction, meaning the total parameter count and theoretical FLOPs are completely identical to the uniform baseline. Yet across three distinct model scales, the Prism Transformer achieves consistent improvements in validation loss and downstream benchmark performance at equivalent training compute budgets. Furthermore, our analysis of per-layer attention distance profiles provides direct mechanistic evidence for this structural optimization, confirming that the network successfully reorganizes its context processing across depth. We view this result as an instance of a broader principle, namely that architectural inductive biases respecting the natural functional specialization of depth can yield genuine representational improvements at no additional cost. The Prism Transformer demonstrates that, in standard autoregressive language models, the head configuration is a meaningful axis of architectural design that has been under-explored.

\paragraph{Limitations}
First, our evaluation is limited to decoder-only transformers for causal language
modeling.  Whether progressive head schedules benefit encoder-only or
encoder-decoder architectures is not evaluated and may differ due to the
bidirectional attention context.  Second, we evaluate at context lengths of
1024 tokens; the impact of the Prism schedule on long-context performance is
left for future work.  Third, our optimal schedule was discovered via a discrete
grid search over a small set of pre-defined configurations; a learned, dynamically adaptive, or
continuous schedule parameterization may yield further improvements.

\bibliographystyle{plain}
\bibliography{references}

\newpage
\appendix




\section{Downstream Benchmark Performance and Scale Dynamics}
\label{sec:appendix_downstream}

We report the complete zero-shot downstream evaluation suite across all three model scales in Table~\ref{tab:benchmarks}. This includes benchmarks evaluating physical common sense (PIQA), common sense reasoning (HellaSwag, WinoGrande), scientific knowledge (ARC-Easy), structural linguistic capability (BLiMP), and language modeling perplexity (WikiText).

\begin{table}[h]
\centering
\caption{
Zero-shot downstream benchmark performance across model scales.
Reported scores are accuracy (\%) and represent the empirical mean
($\mu$) over three independent pre-training seeds.
Bold indicates the best-performing configuration at each scale.
}
\label{tab:benchmarks}
\setlength{\tabcolsep}{5pt}
\renewcommand{\arraystretch}{1.15}
\begin{adjustbox}{max width=\linewidth}
\begin{tabular}{llcccccccc}
\toprule
\textbf{Scale} &
\textbf{Model} &
\textbf{HellaSwag} $\uparrow$ &
\textbf{PIQA} $\uparrow$ &
\textbf{WinoGrande} $\uparrow$ &
\textbf{BLiMP} $\uparrow$ &

\textbf{ARC-E} $\uparrow$ &
\textbf{WikiText} $\downarrow$ \\
\midrule

\multirow{2}{*}{Small}
& Baseline
 & 29.6 & 62.5 & 49.4 & 79.5 & 37.6 & 52.3 \\
 
& \cellcolor{gray!15}\textbf{Prism (Ours)}
 & \cellcolor{gray!15}\textbf{30.0} & \cellcolor{gray!15}\textbf{62.8}
& \cellcolor{gray!15}\textbf{50.2} & \cellcolor{gray!15}\textbf{80.0 }
& \cellcolor{gray!15}\textbf{37.7} & \cellcolor{gray!15}\textbf{51.2} \\
\midrule

\multirow{2}{*}{Medium}
& Baseline
 & 38.7 & 67.7 & 50.2 & \textbf{80.8}  & 42.1 & 33.0 \\
& \cellcolor{gray!15}\textbf{Prism (Ours)}
 & \cellcolor{gray!15}\textbf{39.0} & \cellcolor{gray!15}\textbf{68.2}
& \cellcolor{gray!15}\textbf{50.9} & \cellcolor{gray!15}80.6 
& \cellcolor{gray!15}\textbf{43.1} & \cellcolor{gray!15}\textbf{32.9} \\
\midrule

\multirow{2}{*}{Large}
& Baseline
 & 46.3 & 70.2 & 53.0 & 81.5 & 45.5 & 25.7 \\
& \cellcolor{gray!15}\textbf{Prism (Ours)}
 & \cellcolor{gray!15}\textbf{46.5} & \cellcolor{gray!15}\textbf{71.3}
& \cellcolor{gray!15}\textbf{53.7} & \cellcolor{gray!15}\textbf{81.5}
& \cellcolor{gray!15}\textbf{45.8} & \cellcolor{gray!15}\textbf{25.5} \\
\bottomrule
\end{tabular}
\end{adjustbox}
\end{table}




\section{Optimal Schedules by Scale}
\label{app:schedules}

Table~\ref{tab:optimal_schedules} provides the complete progressive head schedules used across all main experiments, detailed as the explicit head count allocated to each successive layer index. For the Medium and Large configurations (24 layers), the optimal schedule boundaries were determined via a discrete grid search over the same structural axes explored in the Small model ablation study (see Appendix~\ref{sec:appendix_ablations}), treating the starting head count, phase durations, and layer boundaries as free variables.

\begin{table}[h]
\centering
\caption{Optimal progressive head schedules used for the Prism Transformer at each scale. The variable $d_h^{(l)}$ represents the corresponding per-head subspace dimension at layer $l$.}
\label{tab:optimal_schedules}
\setlength{\tabcolsep}{6pt}
\renewcommand{\arraystretch}{1.3}
\small
\begin{tabular}{lcccl}
\toprule
\textbf{Scale} & $d_{\text{model}}$ & $L$ & $h_{\text{base}}$ & \textbf{Prism Schedule (heads per layer)} \\
\midrule
Small  & 768  & 12 & 12 &
    $(\underbrace{3,3}_{},\; \underbrace{6,6}_{},\; \underbrace{8,8}_{},\;
     \underbrace{12,12,12,12,12,12}_{})$ \\[3pt]
Medium & 1024 & 24 & 16 &
    $(\underbrace{4,4,4,}_{},\; \underbrace{8,8,8,}_{},\;
     \underbrace{16,\ldots,16}_{18\times})$ \\[3pt]
Large  & 1536 & 24 & 16 &
    $(\underbrace{6,6,6}_{},\; \underbrace{12,12,12}_{},\;
     \underbrace{16,\ldots,16}_{18\times})$ \\
\bottomrule
\end{tabular}
\end{table}

Note that head counts are mathematically chosen to evenly divide $d_{\text{model}}$ at each phase to satisfy hardware alignment and tensor-parallel partitioning requirements. For the Small configuration ($d_{\text{model}} = 768$), the valid head counts include $\{3, 4, 6, 8, 12\}$, from which we select the progressive sequence $\{3, 6, 8, 12\}$. 

For the larger configurations where $d_{\text{model}} \in \{1024, 1536\}$, the initial head counts are specifically chosen to enforce a uniform initial head dimension invariant of exactly $d_h = 256$ across all model scales ($768/3 = 256$, $1024/4 = 256$, and $1536/6 = 256$). To satisfy this invariant while matching the respective baseline head counts ($h_{\text{base}} = 16$) in the deep layers, the Medium configuration scales through the power-of-two sequence $\{4, 8, 16\}$, while the Large configuration utilizes the sequence $\{6, 12, 16\}$. This progression ensures optimal execution alignment and matrix-tiling factors across distributed hardware layouts while preserving structural parity across model scales.

\section{Ablation Studies}
\label{sec:appendix_ablations}

\subsection{Schedule Sensitivity at Small Scale}
We evaluate eight candidate head schedule configurations at the Small scale (124M parameters, 12 layers) to systematically isolate the architectural properties that dictate representation quality and hardware efficiency. All configurations are pre-trained from scratch using identical compute budgets, optimization hyperparameters, and total parameter allocations as the primary experiments. Table~\ref{tab:ablation_schedules} summarizes these structural variants along three primary architectural axes:
\begin{itemize}
    \item \textbf{Initial Subspace Dimension ($h_1$):} The starting head count in the first layer ($h_1 \in \{2, 3, 4, 6\}$), which controls the maximum initial representational subspace width.
    \item \textbf{Phase Transition Cadence:} The number of unique head count stages (ranging from 2 to 5 phases) deployed across depth prior to arriving at the baseline configuration ($h_{\text{base}} = 12$).
    \item \textbf{Monotonicity and Trajectory:} The direction and smoothness of the head allocation ramp across layers, isolating progressive structures from non-monotonic variations.
\end{itemize}

\begin{table}[h]
\centering
\caption{Ablation study over progressive head schedule configurations at the Small scale ($L=12$, $d_{\text{model}}=768$, $h_{\text{base}}=12$). All variants are trained with identical compute budgets and parameter counts. Reported validation losses and corresponding deltas ($\Delta$) represent the empirical mean across three independent training seeds, calculated relative to the uniform baseline. Throughput changes are evaluated on an $8\times$ H100 GPU cluster.}
\label{tab:ablation_schedules}
\resizebox{\columnwidth}{!}{%
\begin{tabular}{llcccc}
\toprule
\textbf{Config} & \textbf{Head Schedule} & \textbf{Description} & \textbf{Val Loss} $\downarrow$ & $\boldsymbol{\Delta}$ \textbf{vs. Baseline} & \textbf{Tok/s} \\
\midrule
Baseline      & $(12,12,12,12,12,12,12,12,12,12,12,12)$ & Uniform                  & 3.3748& $-$     & $-$ \\
Config-1      & $(3,6,8,12,12,12,12,12,12,12,12,12)$     &  Linear-like Ramp         &3.3691 &     0.0056    & \textcolor[rgb]{0.0, 0.6, 0.0}{$0.0\%$}\\
Config-2      & $(6,6,8,8,12,12,12,12,12,12,12,12)$       & Higher $h_{\text{start}}$         & 3.3683 &    0.0065     & \textcolor[rgb]{0.0, 0.6, 0.0}{$0.0\%$}\\
Config-3      & $(4,4,8,8,12,12,12,12,12,12,12,12)$      & Higher $h_{\text{start}}$                &3.3655 &    0.0093     & \textcolor[rgb]{0.0, 0.6, 0.0}{$0.0\%$}\\

Config-4      & $(6,6,6,6,12,12,12,12,12,12,12,12)$     & 2-phase &3.3647 &  0.0100       & \textcolor[rgb]{0.0, 0.6, 0.0}{$0.0\%$}\\
Config-5      & $(2,4,6,8,12,12,12,12,12,12,12,12)$     & Linear-like Ramp         &3.3647 &    0.0101     & \textcolor[rgb]{0.8, 0.0, 0.0}{$-4.3\%$}\\
Config-6      & $(3,3,6,6,12,12,12,12,12,12,12,12)$         & 3-phase     &3.3628 & 0.0120        & \textcolor[rgb]{0.0, 0.6, 0.0}{$0.0\%$}\\
\textbf{Prism (Ours)} & $\mathbf{(3,3,6,6,8,8,12,12,12,12,12,12)}$ & \textbf{4-phase smooth} & 3.3604 & 0.0143 & \textcolor[rgb]{0.0, 0.6, 0.0}{$0.0\%$}\\
Config-7      & $(2,2,4,4,8,8,12,12,12,12,12,12)$       & Smooth Ramp     & 3.3598&    0.0150     & \textcolor[rgb]{0.8, 0.0, 0.0}{$-8.3\%$}\\
\bottomrule
\end{tabular}%
}
\end{table}

\paragraph{Key Insights and Structural Properties}

\begin{enumerate}
    \item \textbf{The Pareto Frontier of Hardware Alignment:} Config-7 achieves the largest raw validation loss reduction ($\Delta = 0.0150$), but it incurs a substantial hardware throughput penalty of $-8.3\%$. This degradation occurs because an initial head count of 2 forces a per-head dimension of $d_h = 384$, which violates the power-of-two matrix-tiling constraints required by optimized CUDA and Triton attention kernels. Similarly, Config-5 shares the same root cause, yielding a -4.3\% throughput penalty. The Prism Transformer resolves this trade-off by selecting an initial head count of 3 ($d_h = 256$), capturing over $95\%$ of the maximum optimization gains ($\Delta = 0.0143$) while preserving throughput neutrality ($0.0\%$).

    \item \textbf{The Primacy of Initial Subspace Width ($h_1$):} Isolating the starting head configurations reveals a clear trend: as the early attention layers are granted wider, more expressive channels, model performance scales monotonically. Comparing the multi-phase variants across identical layer budgets demonstrates that a starting head count of $h_1 = 6$ (Config-2, $\Delta = 0.0065$) is consistently outperformed by $h_1 = 4$ (Config-3, $\Delta = 0.0093$), which is in turn outperformed by $h_1 = 3$ (Config-6, $\Delta = 0.0120$). This behavior confirms our core architectural premise that maximizing representational bandwidth in early layers yields superior processing efficiency.
    
    \item \textbf{Staircase Smoothness Dictates Representation Stability:} Abrupt structural shifts underperform smoother transitions across depth. For example, transitioning aggressively from wide heads to baseline dimensions in a single step (Config 1 and Config-5) yields a lower improvement than smoother transitions (Prism and Config-7). Sustaining stable head dimensions across consecutive layer blocks, as seen in the multi-phase layout of the Prism Transformer, provides sufficient depth for the network to steadily compose features at a given representational scale before transitioning to finer attention subspaces.


\end{enumerate}

\section{Implementation Details}
\label{app:implementation}

\subsection{Head Schedule in PyTorch}
\label{app:impl_code}

The Prism Transformer requires only a single structural change to a standard
transformer block: the number of attention heads $h_l$ is made
\emph{layer-dependent}, drawn from a fixed schedule
\texttt{n\_heads\_per\_layer}.  All weight matrices retain their original
$\dmodel \times \dmodel$ shape; only the reshape stride in the multi-head
split changes per layer.

Algorithm~\ref{alg:prism_mha} gives the forward pass for a single Prism
attention layer.  Because the head dimension $\dhead = \dmodel / h_l$ varies
across layers, the rotary-embedding buffers ($\cos$, $\sin$) are shared via a
cache keyed on $\dhead$, so each unique head dimension allocates its buffers
exactly once regardless of how many layers share that dimension.

\begin{algorithm}[h]
\caption{Prism Transformer Attention Forward Pass (layer $l$)}
\label{alg:prism_mha}
\begin{algorithmic}[1]
\Require Input $X \in \mathbb{R}^{B \times T \times \dmodel}$;
         fused projection $W_{QKV} \in \mathbb{R}^{\dmodel \times 3\dmodel}$
         (i.e.\ $[W_Q \mid W_K \mid W_V]$ concatenated column-wise),
         output projection $W_O \in \mathbb{R}^{\dmodel \times \dmodel}$;
         head count $h_l$ from the layer schedule;
         shared rotary cache $\mathcal{R}$
\Ensure  Output $Z \in \mathbb{R}^{B \times T \times \dmodel}$
\State $\dhead \gets \dmodel / h_l$
       \Comment{per-layer head dimension}
\State $[Q \mid K \mid V] \gets X W_{QKV}$
       \Comment{single fused linear, shape $(B,\,T,\,3\dmodel)$}
\State Reshape $Q,K,V$ to $(B,\,T,\,h_l,\,\dhead)$
\State $\text{rotary} \gets \mathcal{R}[\dhead]$
       \Comment{select cached \texttt{Rotary} module for this $\dhead$}
\State $(\cos,\sin) \gets \text{rotary}(Q)$
       \Comment{compute RoPE buffers from sequence length $T$}
\State $Q \gets \operatorname{RoPE}(Q,\cos,\sin)$;\quad
       $K \gets \operatorname{RoPE}(K,\cos,\sin)$
       \Comment{$V$ is not positionally encoded}
\State Transpose $Q,K,V$ to $(B,\,h_l,\,T,\,\dhead)$
       \Comment{$V$ unchanged by RoPE}
\State $Y \gets \operatorname{FlashAttn}(Q,\,K,\,V,\;\text{causal}=\top)$
       \Comment{\texttt{scaled\_dot\_product\_attention}}
\State $Y \gets \operatorname{MergeHeads}(Y)$
\Comment{$(B, h_l, T, \dhead) \to (B, T, \dmodel)$}
\State $Z \gets Y W_O$
\State \Return $Z$
\end{algorithmic}
\end{algorithm}




\begin{figure}[h]
    \centering
    \begin{subfigure}{\linewidth}
        \centering
        \includegraphics[width=\linewidth,
            trim={0 0 0 1cm},
    clip
    ]{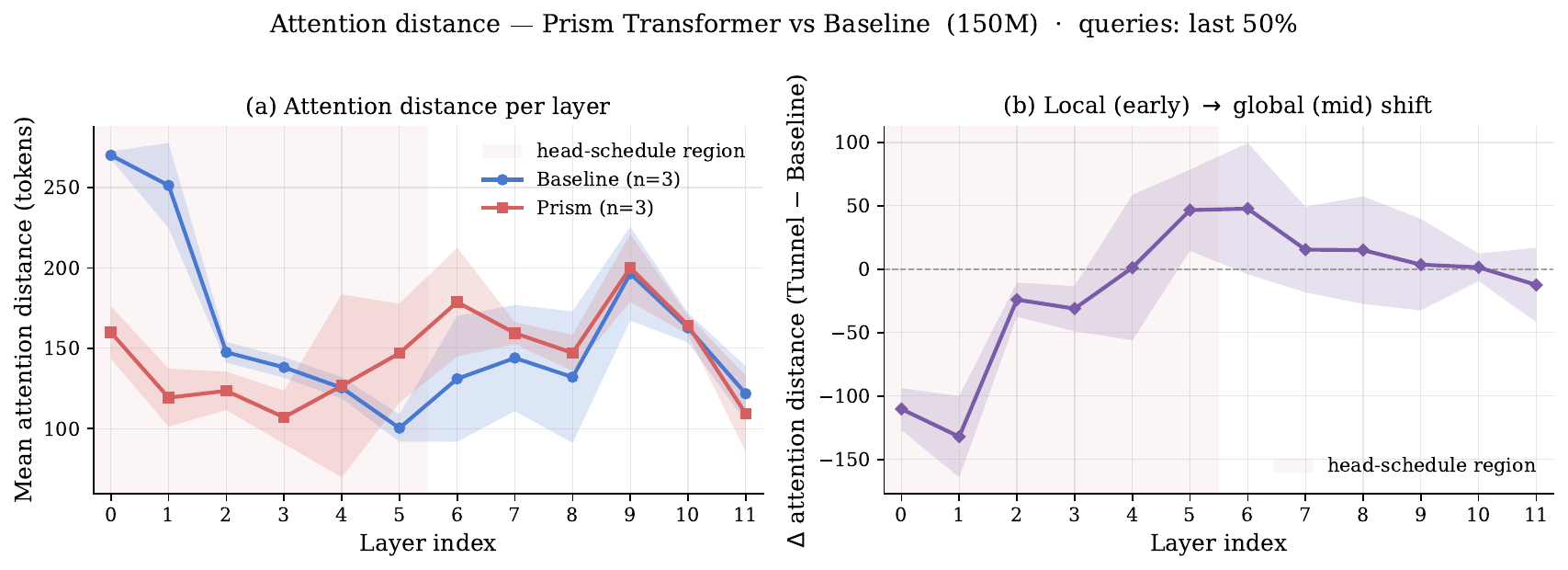}
        \caption{Small (124M).}
        \label{fig:ad_small}
    \end{subfigure}
    \\[0.6em]
    \begin{subfigure}{\linewidth}
        \centering
        \includegraphics[width=\linewidth,
            trim={0 0 0 1cm},
    clip]{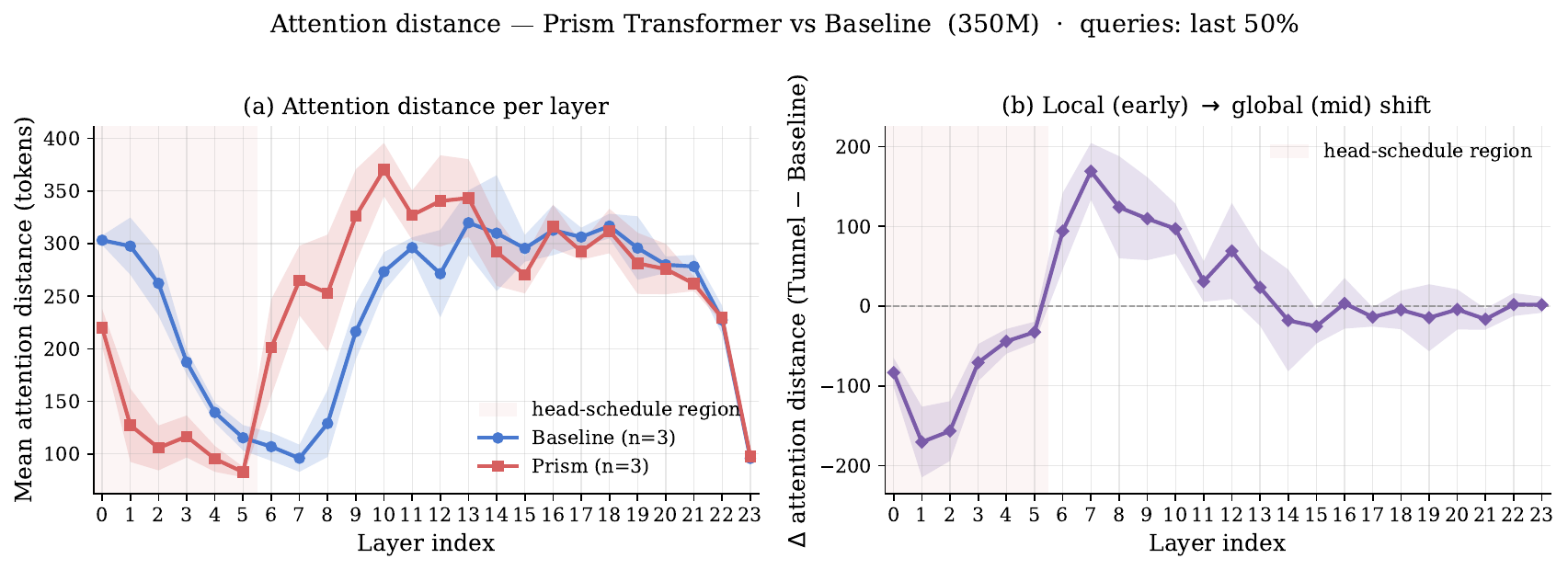}
        \caption{Medium (354M).}
        \label{fig:ad_medium}
    \end{subfigure}
    \\[0.6em]
    \begin{subfigure}{\linewidth}
        \centering
        \includegraphics[width=\linewidth,
            trim={0 0 0 1cm},
    clip]{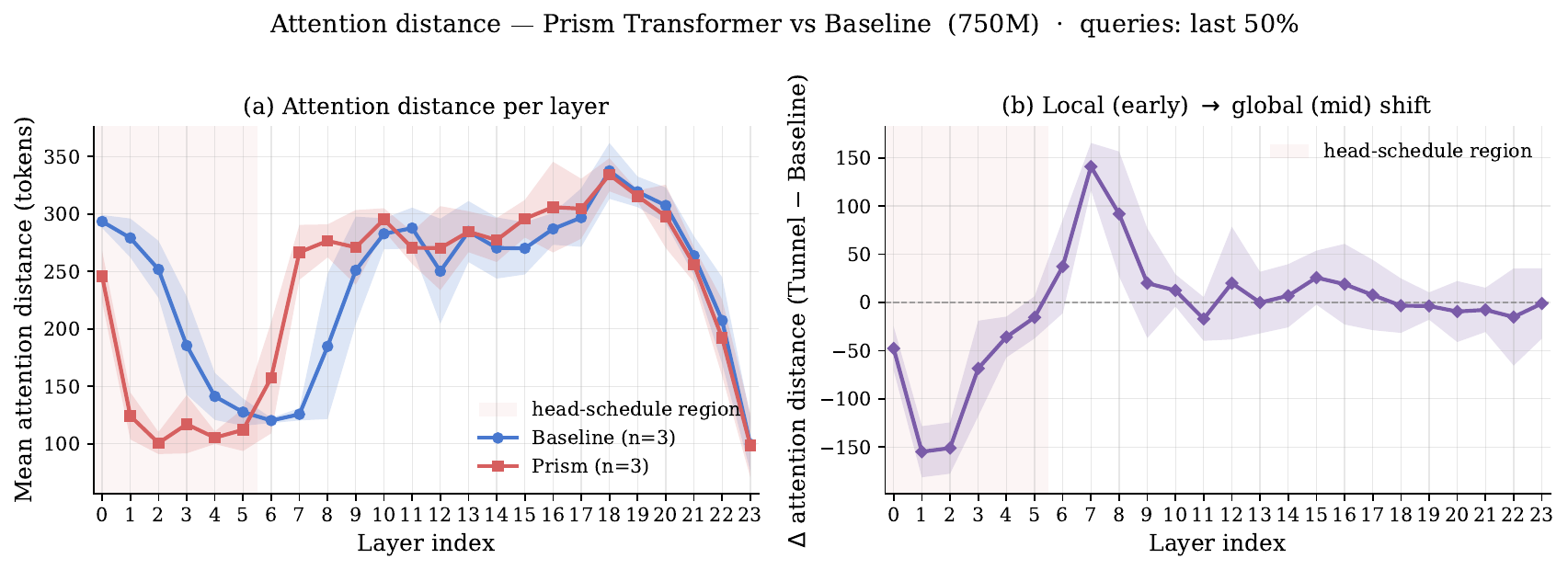}
        \caption{Large (757M).}
        \label{fig:ad_large}
    \end{subfigure}
    \caption{%
        \textbf{Per-layer attention distance, Prism vs.\ Baseline, across three
        scales.}
        For each scale: \emph{(left)} mean attention distance per layer with
        $\pm 1$ s.d.\ bands over $n{=}3$ seeds; \emph{(right)}
        $\Delta\mathcal{D}_l=\mathcal{D}_l^{\text{Prism}}-\mathcal{D}_l^{\text{Base}}$
        with the propagated seed band. The shaded span marks the layers where the
        two head schedules differ (the progressive phase). Relative to the
        baseline, Prism attends \emph{more locally} in the early (wide-head)
        layers and \emph{more globally} in the mid layers where the schedule
        completes, producing a clear sign change in $\Delta\mathcal{D}_l$. The two
        models converge in the late layers. The pattern is consistent across all
        three scales.
    }
    \label{fig:attn_distance}
\end{figure}

\newpage

\end{document}